\documentclass{article}

    \PassOptionsToPackage{numbers}{natbib}

\PassOptionsToPackage{numbers}{natbib}

     \usepackage[preprint]{neurips_2019}

\usepackage[utf8]{inputenc} 
\usepackage[T1]{fontenc}    
\usepackage{hyperref}       
\usepackage{url}            
\usepackage{booktabs}       
\usepackage{amsfonts}       
\usepackage{nicefrac}       
\usepackage{microtype}      
\usepackage{graphicx}
\usepackage{multirow}
\usepackage{subfigure}
\usepackage{caption}
\usepackage{amsmath}

\newcommand{\be}{\begin{equation}}
\newcommand{\ee}{\end{equation}}
\newcommand{\mb}{\mathbf}

\newcommand{\mc}{\mathcal}

\title{Probabilistic Modeling of Deep Features for Out-of-Distribution and Adversarial Detection}

\author{
  Nilesh. A. Ahuja \\ 
  Intel Labs\\
  \And
  Ibrahima Ndiour \\
  Intel Labs \\
  \And
  Trushant Kalyanpur \\
  \And
  Omesh Tickoo \\
  Intel Labs \\
}

\begin{document}
\maketitle

\begin{abstract}
  We present a principled approach for detecting out-of-distribution (OOD) and adversarial samples in deep neural networks. Our approach consists in modeling the outputs of the various layers (deep features) with parametric probability distributions once training is completed. At inference, the likelihoods of the deep features w.r.t the previously learnt distributions are calculated and used to derive uncertainty estimates that can discriminate in-distribution samples from OOD samples.
  We explore the use of two classes of multivariate distributions for modeling the deep features - Gaussian and Gaussian mixture - and study the trade-off between accuracy and computational complexity.  
  We demonstrate benefits of our approach on image features by detecting OOD images and adversarially-generated images, using popular DNN architectures on MNIST and CIFAR10 datasets. We show that more precise modeling of the feature distributions result in significantly improved detection of OOD and adversarial samples; up to 12 percentage points in AUPR and AUROC metrics. 
  We further show that our approach remains extremely effective when applied to video data and associated spatio-temporal features by detecting adversarial samples on activity classification tasks using UCF101 dataset, and the C3D network. To our knowledge, our methodology is the first one reported for reliably detecting white-box adversarial framing, a state-of-the-art adversarial attack for video classifiers.
\end{abstract}

\section{Introduction}
Deep neural networks (DNN) have gained widespread popularity in the last decade, starting with the winning of ILSVRC-2010 challenge by AlexNet \citep{krizhevsky2012imagenet}. Since then, research in this area has led to a proliferation of novelties in methods and architectures that have resulted in dramatic improvements in accuracy \cite{vgg16,He2016DeepRL} and scalability \citep{NIPS2016_6165}. An important area of active research is the ability of DNNs to estimate predictive uncertainty measures, which quantify how much trust should be put in DNN results. This is a critical requirement for perceptual sub-systems based on deep learning, if we are to build safe and transparent systems that do not adversely impact humans (e.g. in fields such as autonomous driving, robotics, or healthcare). 
Additional imperatives for estimating predictive uncertainty measures relate to results interpretability, dataset bias, AI safety, and active learning.

Typically, deep networks do not provide reliable confidence scores for their outputs. Softmax is the most popular score used. Interpreted as a probability, it is a posterior probability and provides a relative ranking of each output with respect to all other outputs, rather than an absolute confidence score. By relying solely on softmax scores as a confidence measure, deep neural networks tend to make overconfident predictions. This is especially true when the input does not resemble the training data (out-of-distribution), or has been crafted to attack and “fool” the network (adversarial examples). 

In this paper, we consider the problem of detecting out-of-distribution (OOD) samples and adversarial samples in DNNs. Recently, there has been substantial work on this topic from researchers in the Bayesian deep learning community \citep{gal2016dropout, kendall2017uncertainties}. In this class of methods, the network's parameters are represented by probability distributions rather than single point values. Such parameters are learned using variational training. At inference, multiple stochastic forward passes are required to generate a distribution over the outputs, instead of the typical single forward pass needed in a traditional (non-Bayesian) DNN. This significantly increases the complexity and requirements in terms of model representation, computational cost and memory. 


Another class of methods attempt to solve this problem by estimating uncertainty directly from a trained DNN (non-Bayesian). \citet{hendrycks2016baseline} proposed using probabilities from the softmax distributions to detect misclassified or OOD samples. \citet{liang2018enhancing} showed that by introducing a temperature scaling parameter to the softmax function, the OOD detection performance could be greatly enhanced relative to \citep{hendrycks2016baseline}. Both these methods use posterior softmax distribution to perform OOD detection. By contrast, \citet{lee2018simple} adopted a generative approach and proposed fitting class-conditional multivariate Gaussian distributions to the pre-trained features of a DNN. The confidence score was defined as the Mahalanobis distance with respect to the closest class conditional distribution. Using this confidence score, they obtained impressive results, outperforming both the previous methods on detecting OOD and adversarial samples. In contrast to Bayesian Deep Learning based approaches, this class of methods can be applied to existing pre-trained networks, does not require weights to be represented by distributions, and does not entail the computational overhead of requiring multiple forward passes during inference.


\paragraph{Contribution:}  
We present an approach for detecting OOD and adversarial samples in deep neural networks based on probabilistic modeling of the deep-features within a DNN. Conceptually, our method is most similar to the generative approach in \citep{lee2018simple} in that we fit class-conditional distributions to the outputs of the various layers (deep-features), once training is completed. In \citep{lee2018simple}, however, it is hypothesized that the class-conditional distributions can be modeled as multivariate Gaussians with shared covariance across all classes. We show that such an assumption is not valid in general; instead, we adopt a more principled approach in choosing the type of density function to model with. To this end, we explore the use of two additional types of distributions to model the deep-features: Gaussian (with separate covariances for each class) and Gaussian mixture models (GMM). We demonstrate that a more precise modeling of the distributions of features results in significantly improved detection of OOD and adversarial samples; in particular, we see an improvement of up to 12 percentage points in the AUPR and AUROC metrics in these tasks. 

We also investigate the numerical and computational aspects of this approach. In particular, fitting distributions to very high-dimensional features can result in severe ill-conditioning during estimation of the densities parameters. We demonstrate empirically that such issues can be resolved with the application of dimensionality reduction techniques such as PCA and average pooling.  

Finally, in addition to demonstrating the effectiveness of the approach on standard image datasets, we further show that our approach remains extremely effective when applied to video data and associated spatio-temporal features by detecting adversarial samples generated by both white-box and black-box attacks on activity classification tasks on the UCF101 dataset \citep{soomro2012ucf101} using the C3D network \citep{hara3dcnns}. To our knowledge, our methodology is the first one reported for reliably detecting white-box adversarial framing \citep{zajac2019framing}, a state-of-the-art adversarial attack for video classifiers. 

\section{Approach}
\label{sec:approach}
\begin{figure}
	\centering
	\includegraphics[width=0.45\textwidth]{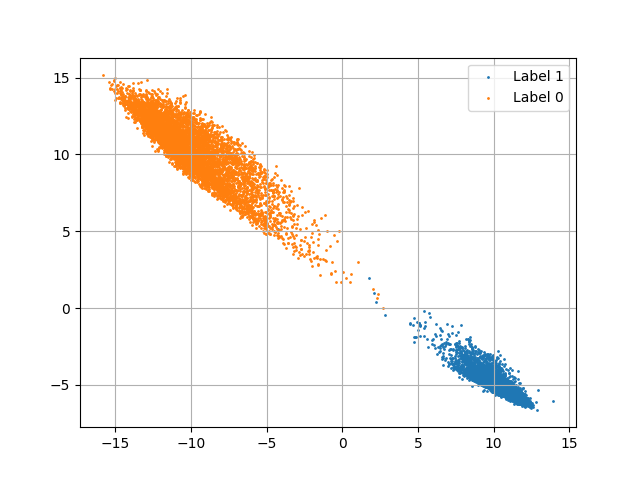}
	\includegraphics[width=0.45\textwidth]{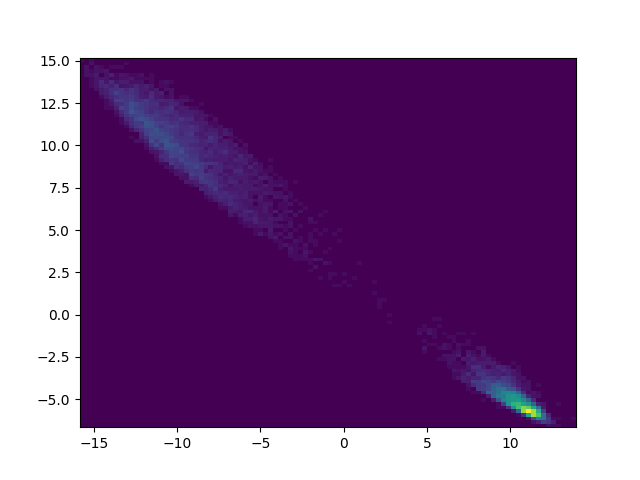}
	\caption{Scatterplot and the corresponding density histogram of the logits. It is clearly seen that the covariances of the two clusters are different.}
	\label{fig:density_histogram_2D}
\end{figure}

Suppose we have a deep network trained to recognize samples from $N$ classes, $\{C_k\}, k=1, \ldots, N$. Let $f_i(\mb{x})$ denote the output at the $i^{th}$ layer of the network, and $n_i$ its dimension. As described earlier, our approach consists of fitting class-conditional probability distributions to the features of a DNN, once training is completed. By fitting distributions to the deep features induced by training samples, we are effectively defining a generative model over the deep feature space. 

At test time, the log-likelihood scores of the features of a test sample are calculated with respect to these distributions and used to derive uncertainty estimates that can discriminate in-distribution samples (which should have high likelihood) from OOD or adversarial  samples (which should have low likelihood). These per-layer likelihoods can also be used for classification in lieu of the softmax output of the network and give classification accuracy as good as the softmax classifier.

{\bf Choice of density function:}
\citet{lee2018simple} assumed that the class-conditional densities $p(f_i(\mb{x})|C_k)$ are multivariate Gaussian with shared covariance across all classes. The justification was based on the following connection between LDA (linear discriminant analysis) and the softmax classifier: in a generative classifier in which the underlying class-conditional distributions are Gaussians with tied covariance, the posterior distribution $p(C_k|f_i(\mb{x}))$ is equivalent to the softmax function with linear separation boundaries \citep{bishop2006pattern}. As we demonstrate empirically via a simple example, the use of a softmax classifier in a DNN does not automatically imply that the underlying distributions will be well represented by a Gaussian with tied covariance. In this example, we constructed and trained a CNN architecture which we call MNET (shown in Figure \ref{fig:mnet}) to classify only two digits ('0' and '1') from the MNIST dataset. Since only two classes are considered, the final FC-10 layer is replaced by FC-2. A 2D density histogram of the features from the FC-2 layer is shown in Figure \ref{fig:density_histogram_2D}. It is obvious even without performing any goodness-of-fit tests that if a 2D Gaussian was fitted to each cluster, the covariance of one would be significantly different from that of the other; forcing these to be the same would result in a poorer fit. Further, even if the assumption of tied-covariance was valid, it would apply only to the features of the final layer of the network, on which softmax is performed. It would not be applicable to the inner layers. 



In this work, therefore, we relax the assumption of tied covariance, and instead employ more general parametric probability distributions. The first type is a separate multivariate Gaussian distribution for each class without the assumption of a tied covariance matrix. Note that this corresponds to the more general QDA (quadratic discriminant analysis) classifier, which is hence capable of representing a larger class of distributions. The second type is a Gaussian Mixture Model (GMM). The choice of GMM is also motivated by the fact that high-dimensional naturally occurring data may not necessarily occupy all dimensions in the Euclidean space, but may in fact reside on or close to an underlying sub-manifold \citep{tenenbaum2000global}. Owing to the more general nature of the GMM, it is a better choice to model such a distribution. In the toy example shown in Figure \ref{fig:gmm-manifold} data is distributed along the boundary of an ellipse. It is clear that the GMM is able to model such a distribution well, while a multivariate Gaussian is a very poor modeling choice for it. It would be interesting to apply more sophisticated manifold learning techniques, but these are significantly more complex to implement practically; their use will be explored in future work. 

{\bf Estimating parameters:} The parameters of the class-conditional densities are estimated from the training set samples by maximum-likelihood. If the chosen density is a multivariate Gaussian (separate covariance), the maximum-likelihood values of the mean and covariance for class $k$ are given by the sample mean and sample covariance:
\be
\label{eq:mean_covar}
\mb{\mu}_k = \frac{1}{M_k}\sum_{\mb{x} \in C_k} f(\mb{\mb{x}}), \quad
\mb{\Sigma}_k = \frac{1}{M_k}\sum_{\mb{x} \in C_k}\left(f(\mb{x})-\mb{\mu}_k)(f(\mb{x})-\mb{\mu}_k\right)^T
\ee
where $f(\mb{x})$ are the feature values from the network and the layer subscript $i$ has been dropped for notational convenience. If the covariance is assumed to be tied across all classes, then all samples $\mb{x}$ in the training set are used to estimate the covariance, rather than only $\mb{x} \in C_K$. The estimation of the mean remains unchanged. If the chosen density is a GMM, its parameters are estimated using an expectation-maximization (EM) procedure. To choose the number of components in the GMM (i.e. model selection), we adopt the Bayesian Information Criteria (BIC) to penalize complex models; details on EM and BIC can be found in \citep{bishop2006pattern}. \\
\begin{minipage}{0.7\textwidth}
	\includegraphics[width=\textwidth]{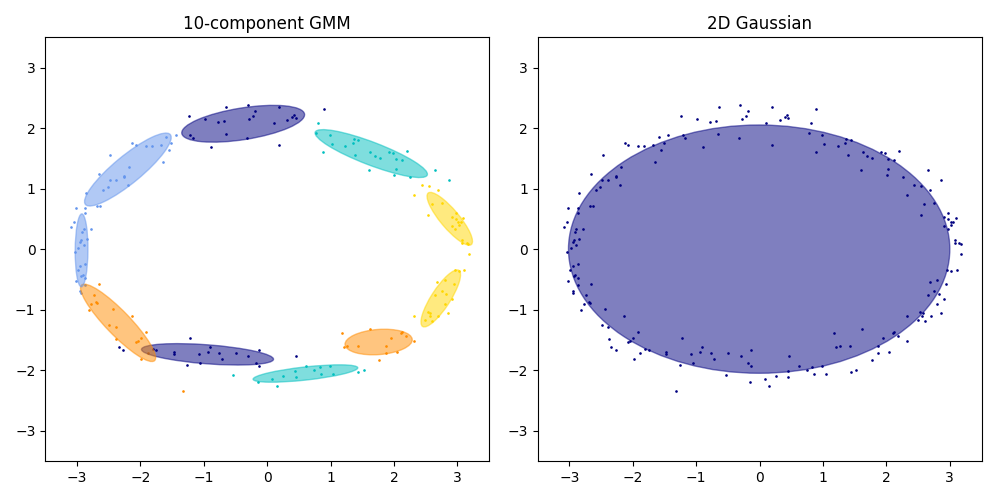}
	\captionof{figure}{Fitting distributions to points on the boundary of an ellipse. The left side has been fitted with a 10-component GMM, is a clearly better fit than the right, which has been fitted with a single 2D Gaussian.}
	\label{fig:gmm-manifold}
\end{minipage}\hspace{1cm}
\begin{minipage}{0.19\textwidth}
  \centering
  \begin{tabular}{|c|}
    \toprule
    input (28 x 28) \\
    \midrule
    conv3-64 \\
    \midrule
    maxpool \\
    \midrule
    conv3-32 \\
    \midrule
    maxpool \\
    \midrule
    FC-128 \\
    \midrule
    FC-10 \\
    \bottomrule
  \end{tabular}
  \captionof{figure}{MNET architecture}
  \label{fig:mnet}
\end{minipage}

{\bf Scoring samples:}
As described earlier, the log-likelihood values are used to measure the closeness of a sample w.r.t a probability distribution. For an $n$-dimensional multivariate Gaussian $\mc{N}(\mb{\mu}, \mb{\Sigma})$, the log-likelihood of a feature $f(\mb{x})$ is given by \citep{bishop2006pattern}:  
\be
2\mc{L} = -\left[n\log2\pi +\log(\det|\mb{\Sigma}|)+(f(\mb{x})-\mb{\mu})^T\Sigma^{-1}(f(\mb{x})-\mb{\mu})\right]
\ee
Under the assumption of tied covariance, the term $n\log2\pi +\log(\det|\mb{\Sigma}|)$ is the same for all the class-conditional distributions and can then be ignored. The remaining term $(f(\mb{x})-\mb{\mu})^T\Sigma^{-1}(f(\mb{x})-\mb{\mu})$, which is the Mahalanobis distance, is then adequate to measure the closeness of a test sample to the modeled distribution. If the covariances are not assumed to be tied, we cannot use the Mahalanobis distance, and should instead use the full log-likelihood term (ignoring the additive and multiplicative constants). For GMM, the log-likelihood is a weighted sum of exponential terms. 

\section{Computational aspects}
\label{sec:computation}
The use of more general distributions such as multivariate Gaussian (without tied covariance) or GMMs to model the class-conditional densities brings its own set of challenges. The obvious one is increased computational complexity, both during modeling and during inference, especially when using GMMs. The other challenge is the lack of sufficient training data from which to estimate the parameters of the modeled distributions. For $n$-dimensional features, if the number of training samples available, $M$, is less than $n$, then the maximum-likelihood estimate of the covariance as given in Eq. (\ref{eq:mean_covar}), will have a rank $M<n$ and be singular. The problem is exacerbated if GMMs are used, since a covariance matrix for each component of each class needs to be estimated, dramatically increasing the number of parameters to be estimated. In such situations, the assumption of tied covariance can prove helpful, since the number of samples used to estimate the covariance matrix would increase by a factor of $N$ (the number of classes), and the covariance matrix can hence be estimated without risk of rank deficiency so long as $mN>n$. However, as we demonstrate, by applying appropriate dimensionality-reduction techniques, we can not only mitigate these issues, but actually improve the eventual detection and classification scores by enabling the use of more general distributions. 

Further, as the dimensionality of the features being modeled increases, it poses numerical challenges which result in highly ill-conditioned covariance matrices. For this reason too, application of some form of dimensionality reduction is recommended. Here, we follow a two-fold approach: average pooling of very high-dimensional layers and applying PCA for projecting onto a lower dimensional subspace. 
In our experiments, we average pool by a factor of 4. This number was chosen empirically, primarily to enhance computational efficiency. While applying PCA, one can specify the fraction of the variance of the original data that should be retained in the lower-dimensional subspace. We choose a high value of 0.995, i.e. we retain 99.5\% of the original variance. This resulted in a dramatic reduction in the feature dimensions, at times up to 90\%, indicating that 99.5\% of the information in the features is actually contained in a much lower- dimensional subspace.

\section{Experiments and Results}
\label{sec:Results}

\subsection{Applications in Image Classification tasks}

\paragraph{Experimental setup and evaluation metrics}
\label{ref:exp_setup}
We use MNIST and CIFAR10 as the in-distribution datasets. For MNIST, we use FashionMNIST and EMNIST Letters \citep{cohen2017emnist} as the OOD datasets. For CIFAR10, we use SVHN dataset \citep{netzer2011reading} and a resized version of the LSUN datasets \citep{yu15lsun} as the OOD datasets. To test against adversarial attacks, we use the FGSM attack introduced by \citet{goodfellow2014explaining}. In all experiments, the parameters of the fitted density function are estimated from the training split of the in-distribution dataset, while performance metrics (accuracy, AUPR, AUROC) are calculated on the test split.

For MNIST, use the MNET architecture as shown in Figure \ref{fig:mnet}. For CIFAR10, we use two publicly available architectures: Resnet50 and Densenet-BC. For reasons of computational efficiency, we perform our experiments on 3 layers of the networks listed above. In MNET, these are the final 3 layers; in Densenet and Resnet, these are the outputs of the corresponding final 3 dense or residual blocks. The layers are labelled as 0, 1, and 2, with 0 being the outermost layer, and 1,2 being inside the network. Layers further inside can easily be included too, but these typically have outputs of very high-dimensions and require aggressive dimensionality reduction in order to process them efficiently.

\begin{table}
  \caption{Classification accuracy}
  \label{table:accuracy}
  \centering
  \begin{tabular}{cccccccccc}
    \toprule
     & \multicolumn{3}{c}{MNIST} & \multicolumn{3}{c}{CIFAR10(Resnet)} & \multicolumn{3}{c}{CIFAR10(densenet)} \\
    \cmidrule(r){2-4}\cmidrule(r){5-7}\cmidrule(r){8-10}
     & GMM & Sep & Tied & GMM & Sep & Tied & GMM & Sep & Tied \\
    \cmidrule(r){1-1}\cmidrule(r){2-4}\cmidrule(r){5-7}\cmidrule(r){8-10}

     Layer 0 & 98.9 & 98.6  & 98.6  & 90.2 & {\bf 90.5} & 89.9 & {\bf 90.3} & 90.0 & 89.8 \\
     Layer 1 & 98.2 & 98.6 & 98.6 & 89.9 & 90.5 & 90.0 & 89.1 & 88.7 & 89.7\\
     Layer 2 & 86.0 & 97.4 & 98.3 & 90.1 &	90.5 & 90.0 & 88.9	& 88.9 & 89.9 \\
    \cmidrule(r){2-4}\cmidrule(r){5-7}\cmidrule(r){8-10}
    Softmax & \multicolumn{3}{c}{\bf 98.99} & \multicolumn{3}{c}{89.09} & \multicolumn{3}{c}{89.14}\\
    \bottomrule
  \end{tabular}
\end{table}

During testing, the log-likelihood scores of the features generated by a test sample are calculated. These can then be used to distinguish between in-distribution and out-of-distribution data, effectively creating a binary classifier. The performance of this classifier can be characterized by typical methods such as the precision-recall (PR) curve or the receiver operating characteristics (ROC curve). \citet{davis2006relationship} showed that although the PR and ROC curves are equivalent, maximizing the area under ROC (AUROC) is \emph{not} equivalent to maximizing area under precision-recall (AUPR). We report, therefore, both metrics in our results. To calculate these metrics, we used the scikit-learn library \citep{scikit-learn}.

\paragraph{Results}
We want to first demonstrate the effectiveness of our approach by using it to perform classification based solely on the log-likelihood scores w.r.t the class-conditional distributions of a particular layer. The classification accuracy using this scheme is measured on the test set for each layer individually. The results are summarized in Table \ref{table:accuracy}. It is seen that the classification accuracy using the proposed method is comparable, if not slightly better, than the softmax-based accuracy, indicating that our scheme is as good as softmax for classification of in-distribution samples.
\begin{table}
  \caption{AUPR (\%) scores from three different density functions: GMM, Sep (Gaussian with separate covariance per class), Tied (Gaussian with tied covariance). Best values are shown in {\bf bold}.}
  \label{table:aupr}
  \centering
  \begin{tabular}{cccccccccc}
    \toprule
    \multirow{2}{*}{MNIST} 
     & \multicolumn{3}{c}{FashionMNIST} & \multicolumn{3}{c}{EMNIST} & \multicolumn{3}{c}{FGSM, $\epsilon=0.2$} \\
    \cmidrule(r){2-4}\cmidrule(r){5-7}\cmidrule(r){8-10}
     & GMM & Sep & Tied & GMM & Sep & Tied & GMM & Sep & Tied \\
    \cmidrule(r){1-1}\cmidrule(r){2-4}\cmidrule(r){5-7}\cmidrule(r){8-10}
	   Layer 0 & {\bf 86.4} & 84.8 & 81.8 & {\bf 66.5} & 61.1 & 53.4 & {\bf 96.0}  & 95.3 & 94.8 \\
       Layer 1 & {\bf 84.5} & 81.3 & 53.5 & 67.5 & {\bf 72.8} & 65.4 & {\bf 96.4} & 95.3 & 88.0 \\
       Layer 2 & 88.4 & {\bf 90.9} & 58.1 & 72.0  & {\bf 77.7} & 58.1 & {\bf 97.7} & 97.7 & 86.1 \\
    \midrule
    \multirow{2}{*}{\begin{tabular}{c}
         CIFAR10  \\ (Resnet)
    \end{tabular}} 
     & \multicolumn{3}{c}{SVHN} & \multicolumn{3}{c}{LSUN} & \multicolumn{3}{c}{FGSM, $\epsilon=0.1$} \\
    \cmidrule(r){2-4}\cmidrule(r){5-7}\cmidrule(r){8-10}
     & GMM & Sep & Tied & GMM & Sep & Tied & GMM & Sep & Tied \\
    \cmidrule(r){1-1}\cmidrule(r){2-4}\cmidrule(r){5-7}\cmidrule(r){8-10}
	   Layer 0 & {\bf 95.0} & 93.8 & 91.5& {\bf 64.4}& 62.8 & 56.0 & {\bf 92.5} & 90.7& 89.8 \\
       Layer 1 & {\bf 95.2} & 94.9 & 93.4 & 57.9 & {\bf 68.8} & 68.0 & {\bf 93.2} & 92.8& 92.4 \\
       Layer 2 & {\bf 95.1} & 94.9 & 93.4 & 57.9 & {\bf 68.8} & 68.0 & {\bf 93.2} & 92.8& 92.4 \\
    \midrule
    \multirow{2}{*}{\begin{tabular}{c}
         CIFAR10  \\ (Densenet)
    \end{tabular}} 
     & \multicolumn{3}{c}{SVHN} & \multicolumn{3}{c}{LSUN} & \multicolumn{3}{c}{FGSM, $\epsilon=0.1$} \\
    \cmidrule(r){2-4}\cmidrule(r){5-7}\cmidrule(r){8-10}
     & GMM & Sep & Tied & GMM & Sep & Tied & GMM & Sep & Tied \\
    \cmidrule(r){1-1}\cmidrule(r){2-4}\cmidrule(r){5-7}\cmidrule(r){8-10}
	   Layer 0 & 79.1 & 77.2 & {\bf 79.6} & {\bf 29.5} & 28.9 & 21.9 & 85.8 & {\bf 86.1} & 84.6 \\
	   Layer 1 & {\bf 86.8} & 85.6 & 75.9 & 77.7 & {\bf 80.4} & 79.3 & {\bf 90.6} & 90.2 & 87.0 \\
	   Layer 2 & {\bf 80.4} & 80.4 & 57.1 & {\bf 37.2} & 37.2 & 27.1 & 87.2 & {\bf 87.2} & 78.1 \\
    \bottomrule
  \end{tabular}
\end{table}
\begin{table}
  \caption{AUROC scores (\%) from three different density functions: GMM, Sep (Gaussian with separate covariance per class), Tied (Gaussian with tied covariance). Best values are shown in {\bf bold}.}
  \label{table:roc}
  \centering
  \begin{tabular}{cccccccccc}
    \toprule
    \multirow{2}{*}{MNIST} 
     & \multicolumn{3}{c}{FashionMNIST} & \multicolumn{3}{c}{EMNIST} & \multicolumn{3}{c}{FGSM, $\epsilon=0.2$} \\
    \cmidrule(r){2-4}\cmidrule(r){5-7}\cmidrule(r){8-10}
     & GMM & Sep & Tied & GMM & Sep & Tied & GMM & Sep & Tied \\
    \cmidrule(r){1-1}\cmidrule(r){2-4}\cmidrule(r){5-7}\cmidrule(r){8-10}
	   Layer 0 & {\bf 92.9} & 91.8 & 92.1 & {\bf 94.0}  & 93.2 & 91.9 & {\bf 87.2} & 85.9 & 84.4 \\
       Layer 1 & 92.9 & {\bf 93.5} & 75.3 & 96.2 & {\bf 96.3} & 93.4 & {\bf 88.6} & 85.7 & 66.8 \\
       Layer 2 & 97.0  & {\bf 97.5} & 89.0  & 96.7 & {\bf 97.2} & 93.1 & {\bf 92.0}  & 92.0  & 62.6 \\
    \midrule
    \multirow{2}{*}{\begin{tabular}{c}
         CIFAR10  \\ (Resnet)
    \end{tabular}} 
     & \multicolumn{3}{c}{SVHN} & \multicolumn{3}{c}{LSUN} & \multicolumn{3}{c}{FGSM, $\epsilon=0.1$} \\
    \cmidrule(r){2-4}\cmidrule(r){5-7}\cmidrule(r){8-10}
     & GMM & Sep & Tied & GMM & Sep & Tied & GMM & Sep & Tied \\
    \cmidrule(r){1-1}\cmidrule(r){2-4}\cmidrule(r){5-7}\cmidrule(r){8-10}
	   Layer 0 & {\bf 94.2} & 93.1 & 90.0  & 78.2 & {\bf 80.8} & 79.8 & {\bf 91.0} & 89.0 & 87.9 \\
       Layer 1 & {\bf 94.3} & 94.3 & 92.3 & 75.3 & 82.9 & {\bf 84.3} & {\bf 92.5} & 91.9 & 91.8 \\
       Layer 2 & 94.2 & {\bf 94.3} & 92.3 & 75.7 & 82.9 & {\bf 84.3} & {\bf 92.4} & 91.9 & 91.8 \\

    \midrule
    \multirow{2}{*}{\begin{tabular}{c}
         CIFAR10  \\ (Densenet)
    \end{tabular}} 
     & \multicolumn{3}{c}{SVHN} & \multicolumn{3}{c}{LSUN} & \multicolumn{3}{c}{FGSM, $\epsilon=0.1$} \\
    \cmidrule(r){2-4}\cmidrule(r){5-7}\cmidrule(r){8-10}
     & GMM & Sep & Tied & GMM & Sep & Tied & GMM & Sep & Tied \\
    \cmidrule(r){1-1}\cmidrule(r){2-4}\cmidrule(r){5-7}\cmidrule(r){8-10}
       Layer 0 & 76.7 & 75.2 & {\bf 77.2} & 69.5 & {\bf 69.9} & 64.8 & 85.7 & {\bf 86.4} & 83.6 \\
       Layer 1 & {\bf 85.2} & 84.7 & 73.3 & 94.8 & 95.2 & {\bf 95.4} & {\bf 92.7} & 92.5 & 90.3 \\
       Layer 2 & {\bf 78.1} & 78.1 & 51.1 & {\bf 78.2} & 78.2 & 74.8 & {\bf 88.0}  & 88.0  & 80.5 \\
    \bottomrule
  \end{tabular}
\end{table}

\begin{table}
  \caption{Average improvements in scores over the baseline distribution}
  \label{table:avg_improv}
  \centering
  \begin{tabular}{ccccc}
    \toprule
     & \multicolumn{2}{c}{AUPR change} & \multicolumn{2}{c}{AUROC change} \\
    \cmidrule(r){2-3}\cmidrule(r){4-5}
     & GMM & Sep & GMM & Sep\\
    \cmidrule(r){2-3}\cmidrule(r){4-5}
    Layer 0 & 4.64 & 3.00 & 1.97 & 1.51 \\
    Layer 1 & 5.21 & 6.60 & 5.51 & 6.01 \\
    Layer 2 & 10.1 & 12.1 & 8.09 &	8.96 \\
    \bottomrule
  \end{tabular}
\end{table}

To see the performance on OOD samples, we calculate the AUPR and AUROC scores as described earlier. In particular, we examine the change in AUPR and AUROC values obtained by using the more general distribution types (outlined in Section \ref{sec:approach}) relative to those obtained by using the baseline distribution (multivariate Gaussian with tied covariance). The results are presented in Tables \ref{table:aupr} and \ref{table:roc}. It is seen that the use of the more general distribution types results in improvements, often significant, in the AUPR and AUROC scores over the baseline distribution. On the few instances in which the baseline distribution achieves the best score, it is by a small margin. It is further interesting to examine the improvements in scores over the baseline distribution as a function of the layer being modeled. These results are summarized in Table \ref{table:avg_improv}, which shows the average change (across all tested datasets) in the AUPR and AUROC scores per layer. For all layers, switching to a more general distribution produces an improvement in the scores. However, the extent of the improvement increases the further we are from the final output layer. This is consistent with the reasoning described in Section \ref{sec:approach} that the assumption of a Gaussian with tied covariance is a valid approximation for the output layer only, and not the inner layers. 

Finally, note that the improvements obtained by using a multivarate Gaussian (separate covariance) and GMM are very comparable. This observation is consistent with the reasoning in Section \ref{sec:computation} that the larger number of parameters to be estimated for a GMM can cause more ill-conditioning during estimation, especially in the case of high-dimension and limited training samples, which might lead to sub-optimal parameter estimates.  

\subsection{Application to Activity Classification in Videos}
While there is extensive work on adversarial attacks against image classifiers, the reported cases of video classifier attacks remains very limited. Here, we consider one such case on a video-based human activity classifier. The setup is the following: we use the original UCF101 dataset (split into training and testing subsets) to train a 3D deep neural network, C3D ResNet101 \citep{hara3dcnns}, for human action classification. With our trained model, we obtain $84.1\%$ accuracy on the test set. We then use a state-of-the-art video attack method known as adversarial framing \cite{zajac2019framing} to generate adversarial samples. Adversarial framing is a simple, yet very potent video attack method that operates by keeping most of the video frames intact, but adds a 4-pixel wide framing at the border of each frame. We employ both a white-box attack where we assume full knowledge of the action recognition classifier (architecture and parameters) and a black-box attack where no such knowledge is available. These allow us to generate two sets of adversarial frames of the UCF101 dataset that are fed as inputs to the action classifier. The classifier's recognition accuracy drops to $63.1\%$ and $4.2\%$ for black-box and white-box attacks respectively. For the sake of brevity, we show visualizations only for the white-box attack, but results for both types of attacks are fully reported in Table \ref{table:video_metrics}.

\begin{figure}
	\centering
	\includegraphics[width=0.49\textwidth]{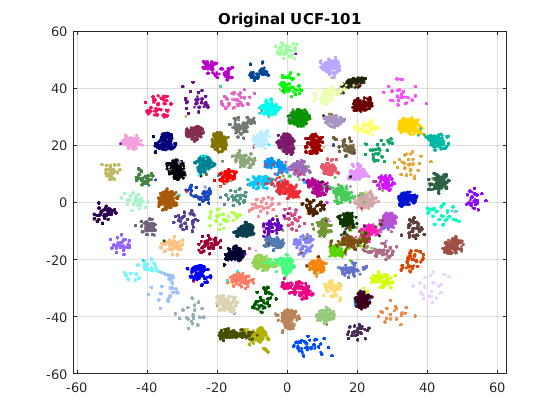}
	\includegraphics[width=0.49\textwidth]{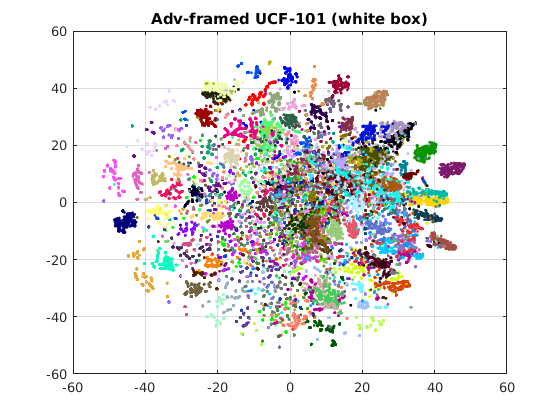}
	\caption{t-SNE visualization of (spatio-temporal) feature embeddings for UCF101 using C3D Resnet101 (Layer 1).}
	\label{fig:tsne}
\end{figure}

\begin{figure}
    \centering
    \subfigure[Softmax]{\includegraphics[width=0.45\textwidth]{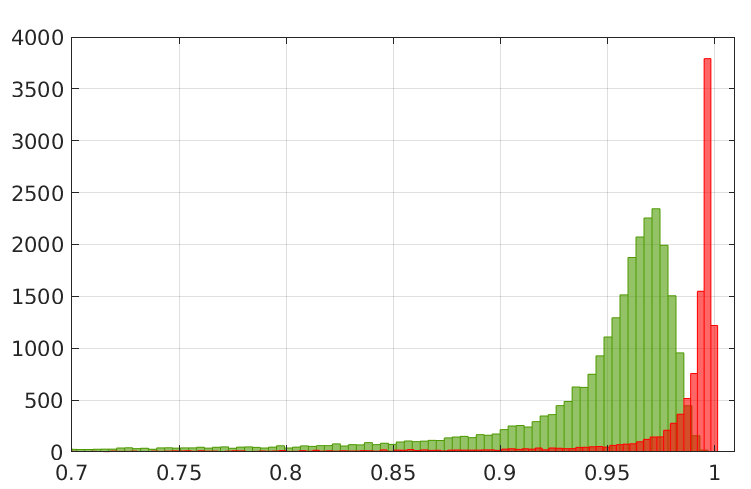}} 
    \subfigure[Gaussian with tied covariance]{\includegraphics[width=0.45\textwidth]{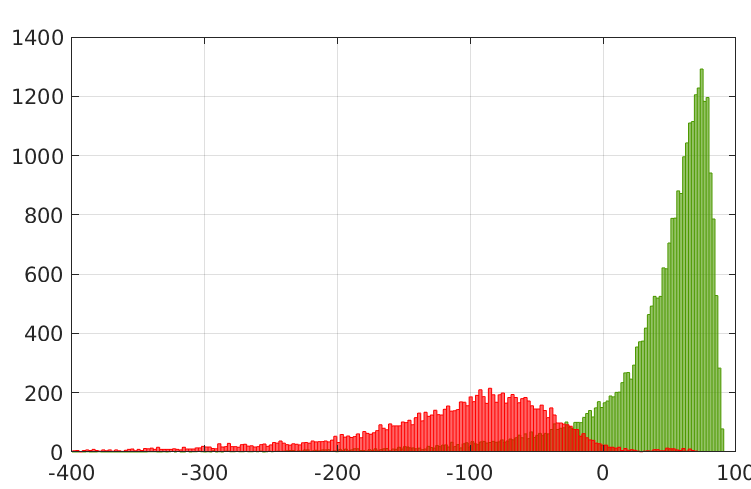}} 
    \subfigure[Gaussian with separate covariances]{\includegraphics[width=0.45\textwidth]{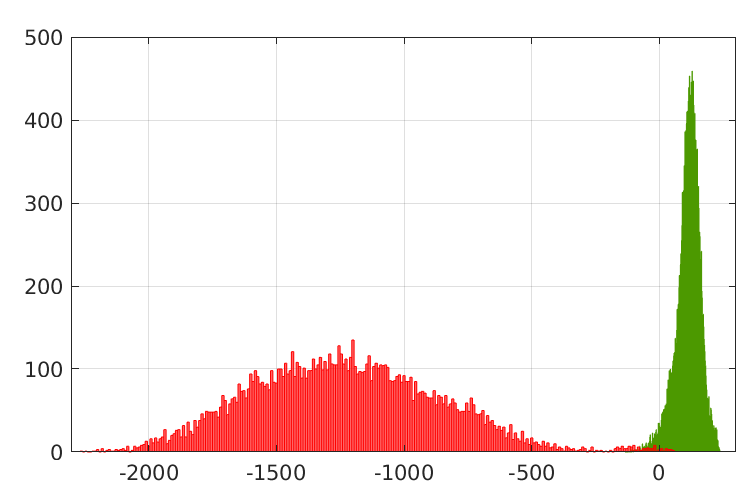}}
    \subfigure[Gaussian mixture]{\includegraphics[width=0.45\textwidth]{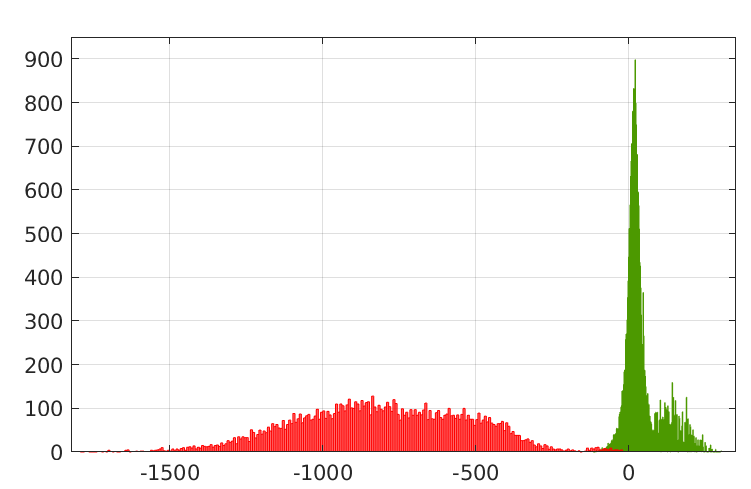}}
    \caption{Density histogram of softmax and log-likelihood scores for Layer 0 (in-distribution samples in green, white-box adversarial samples in red).}
    \label{fig:density_scores}
\end{figure}

For this experiment, we fit distributions to the features from the logit layer (output of last layer before softmax) and the preceding layer (feature embeddings), denoted as Layers 0 and 1 respectively. Figure \ref{fig:tsne} provides a visualization of the feature embeddings via t-SNE \citep{maaten2008visualizing} for the genuine data and white-box attacked data, showing the potency of the white-box adversarial framing attack. Subsequently, the adversarial samples are passed through the network and the corresponding uncertainty (log-likelihood) scores at layers 1 and 0 are calculated. Figure \ref{fig:density_scores} shows the density histogram of these scores for the in-distribution data and white-box attacked data. Note that while the recognition accuracy dramatically plummeted to $4.2\%$ for the white-box adversarial samples, the softmax scores shows even more confidence: the network outputs are wrong, yet the network is ever more confident. Using our approach, Figure \ref{fig:density_scores} shows that while the network still provides wrong classification results, it is now able to produce reliable uncertainty metrics showing poor confidence in the generated outputs. Moreover, the discrimination between in-distribution and OOD samples (adversarial here) is clearly improved with the more general choices of distributions (GMM and Gaussian with separate covariances). These improvements are captured quantitatively with AUPR and AUROC metrics in Table \ref{table:video_metrics} for both white-box and black-box attacks.

\begin{table}
  \caption{Quantitative metrics (AUROC, AUPR) for video attack detection}
  \label{table:video_metrics}
  \centering
  \begin{tabular}{cp{0.58cm}p{0.58cm}p{0.58cm}p{0.58cm}p{0.58cm}p{0.58cm}p{0.58cm}p{0.58cm}p{0.58cm}p{0.58cm}p{0.58cm}p{0.58cm}}
    \toprule
     & \multicolumn{6}{c}{White box attack} & \multicolumn{6}{c}{Black box attack} \\
     & \multicolumn{3}{c}{AUROC} & \multicolumn{3}{c}{AUPR} & \multicolumn{3}{c}{AUROC} & \multicolumn{3}{c}{AUPR}  \\
    \cmidrule(r){2-4}\cmidrule(r){5-7}\cmidrule(r){8-10}\cmidrule(r){11-13}
     & GMM & Sep & Tied & GMM & Sep & Tied & GMM & Sep & Tied & GMM & Sep & Tied\\
    \cmidrule(r){1-1}\cmidrule(r){2-4}\cmidrule(r){5-7}\cmidrule(r){8-10}\cmidrule(r){11-13}

     Layer 0 & 97.6 & {\bf 99.4}  & 83.2  & 97.7 & {\bf 99.3} & 88.4 & 86.6 & {\bf 90.8} & 83.6 & 90.3 & {\bf 93.2} & 87.6 \\
     Layer 1 & 91.3 & {\bf 92.3} & 91.7 & 92.9 &  93.7 &  {\bf 93.9} & 86.4 &  {\bf 86.9} & 83.6 & 90.4 & {\bf 90.8} & 87.3\\
    \cmidrule(r){1-1}\cmidrule(r){2-4}\cmidrule(r){5-7}\cmidrule(r){8-10}\cmidrule(r){11-13}
    Softmax & \multicolumn{3}{c}{12.0} & \multicolumn{3}{c}{32.8} & \multicolumn{3}{c}{72.7} & \multicolumn{3}{c}{76.8} \\
    \bottomrule
  \end{tabular}
\end{table}


\section{Conclusions and Future Work}
This paper presented a method for modeling the outputs of the various DNN layers (deep-features) with parametric probability distributions, with applications to adversarial and out-of-distribution sample detection. We showed that accurate modeling of the class-conditional distributions can enable the derivation of reliable uncertainty scores. The methodology was theoretically motivated, and experimentally proven by showing improvements to out-of-distribution detection, and adversarial sample detection on both image and video data. In particular, we report adversarial sample detection against a state-of-the-art video classifier attack.

While this work performed feature modeling based on a trained model, future work will seek to analyze the evolution of the feature distributions during training. Given the complexities arising from parameter estimation on high-dimensional spaces, we will also consider fitting distributions on features induced by larger pre-training datasets (e.g. ImageNet, Sports1M, Kinetics \cite{CaZi2017Kinetics}) and subsequently use the estimated parameters as priors for modeling the features on the (smaller) dataset of interest.

\bibliographystyle{apalike}
\bibliography{neurips_2019}

\begin{thebibliography}{}

\bibitem[Bishop, 2006]{bishop2006pattern}
Bishop, C.~M. (2006).
\newblock {\em Pattern recognition and machine learning}.
\newblock springer.

\bibitem[Carreira and Zisserman, 2017]{CaZi2017Kinetics}
Carreira, J. and Zisserman, A. (2017).
\newblock Quo vadis, action recognition? a new model and the kinetics dataset.
\newblock In {\em Proceedings of the IEEE Conference on Computer Vision and
  Pattern Recognition (CVPR)}, page 4724–4733.

\bibitem[Cohen et~al., 2017]{cohen2017emnist}
Cohen, G., Afshar, S., Tapson, J., and van Schaik, A. (2017).
\newblock Emnist: an extension of mnist to handwritten letters.
\newblock {\em arXiv preprint arXiv:1702.05373}.

\bibitem[Davis and Goadrich, 2006]{davis2006relationship}
Davis, J. and Goadrich, M. (2006).
\newblock The relationship between precision-recall and roc curves.
\newblock In {\em Proceedings of the 23rd international conference on Machine
  learning}, pages 233--240. ACM.

\bibitem[Gal and Ghahramani, 2016]{gal2016dropout}
Gal, Y. and Ghahramani, Z. (2016).
\newblock Dropout as a bayesian approximation: Representing model uncertainty
  in deep learning.
\newblock In {\em international conference on machine learning}, pages
  1050--1059.

\bibitem[Goodfellow et~al., 2015]{goodfellow2014explaining}
Goodfellow, I.~J., Shlens, J., and Szegedy, C. (2015).
\newblock Explaining and harnessing adversarial examples.

\bibitem[Guo et~al., 2016]{NIPS2016_6165}
Guo, Y., Yao, A., and Chen, Y. (2016).
\newblock Dynamic network surgery for efficient dnns.
\newblock In {\em Advances in Neural Information Processing Systems}, pages
  1379--1387.

\bibitem[Hara et~al., 2018]{hara3dcnns}
Hara, K., Kataoka, H., and Satoh, Y. (2018).
\newblock Can spatiotemporal 3d cnns retrace the history of 2d cnns and
  imagenet?
\newblock In {\em Proceedings of the IEEE Conference on Computer Vision and
  Pattern Recognition (CVPR)}, pages 6546--6555.

\bibitem[He et~al., 2016]{He2016DeepRL}
He, K., Zhang, X., Ren, S., and Sun, J. (2016).
\newblock Deep residual learning for image recognition.
\newblock In {\em Proceedings of the IEEE Conference on Computer Vision and
  Pattern Recognition (CVPR)}, pages 770--778.

\bibitem[Hendrycks and Gimpel, 2017]{hendrycks2016baseline}
Hendrycks, D. and Gimpel, K. (2017).
\newblock A baseline for detecting misclassified and out-of-distribution
  examples in neural networks.

\bibitem[Kendall and Gal, 2017]{kendall2017uncertainties}
Kendall, A. and Gal, Y. (2017).
\newblock What uncertainties do we need in bayesian deep learning for computer
  vision?
\newblock In {\em Advances in neural information processing systems}, pages
  5574--5584.

\bibitem[Krizhevsky et~al., 2012]{krizhevsky2012imagenet}
Krizhevsky, A., Sutskever, I., and Hinton, G.~E. (2012).
\newblock Imagenet classification with deep convolutional neural networks.
\newblock In {\em Advances in neural information processing systems}, pages
  1097--1105.

\bibitem[Lee et~al., 2018]{lee2018simple}
Lee, K., Lee, K., Lee, H., and Shin, J. (2018).
\newblock A simple unified framework for detecting out-of-distribution samples
  and adversarial attacks.
\newblock In {\em Advances in Neural Information Processing Systems}, pages
  7167--7177.

\bibitem[Liang et~al., 2018]{liang2018enhancing}
Liang, S., Li, Y., and Srikant, R. (2018).
\newblock Enhancing the reliability of out-of-distribution image detection in
  neural networks.

\bibitem[Maaten and Hinton, 2008]{maaten2008visualizing}
Maaten, L. v.~d. and Hinton, G. (2008).
\newblock Visualizing data using t-sne.
\newblock {\em Journal of machine learning research}, 9(Nov):2579--2605.

\bibitem[Netzer et~al., 2011]{netzer2011reading}
Netzer, Y., Wang, T., Coates, A., Bissacco, A., Wu, B., and Ng, A.~Y. (2011).
\newblock Reading digits in natural images with unsupervised feature learning.

\bibitem[Pedregosa et~al., 2011]{scikit-learn}
Pedregosa, F., Varoquaux, G., Gramfort, A., Michel, V., Thirion, B., Grisel,
  O., Blondel, M., Prettenhofer, P., Weiss, R., Dubourg, V., Vanderplas, J.,
  Passos, A., Cournapeau, D., Brucher, M., Perrot, M., and Duchesnay, E.
  (2011).
\newblock Scikit-learn: Machine learning in {P}ython.
\newblock {\em Journal of Machine Learning Research}, 12:2825--2830.

\bibitem[Simonyan and Zisserman, 2015]{vgg16}
Simonyan, K. and Zisserman, A. (2015).
\newblock Very deep convolutional networks for large-scale image recognition.

\bibitem[Soomro et~al., 2012]{soomro2012ucf101}
Soomro, K., Zamir, A.~R., and Shah, M. (2012).
\newblock Ucf101: A dataset of 101 human actions classes from videos in the
  wild.
\newblock {\em arXiv preprint arXiv:1212.0402}.

\bibitem[Tenenbaum et~al., 2000]{tenenbaum2000global}
Tenenbaum, J.~B., De~Silva, V., and Langford, J.~C. (2000).
\newblock A global geometric framework for nonlinear dimensionality reduction.
\newblock {\em science}, 290(5500):2319--2323.

\bibitem[Yu et~al., 2015]{yu15lsun}
Yu, F., Zhang, Y., Song, S., Seff, A., and Xiao, J. (2015).
\newblock Lsun: Construction of a large-scale image dataset using deep learning
  with humans in the loop.
\newblock {\em arXiv preprint arXiv:1506.03365}.

\bibitem[Zaj\k{a}c et~al., 2019]{zajac2019framing}
Zaj\k{a}c, M., \.Zo\l{}na, K., Rostamzadeh, N., and Pinheiro, P. (2019).
\newblock Adversarial framing for image and video classification.
\newblock In {\em AAAI Conference on Artificial Intelligence}.

\end{thebibliography}

\end{document}